# Self-reconfiguration Strategies for Space-distributed Spacecraft

Tianle Liu, Zhixiang Wang, Yongwei Zhang, Ziwei Wang, Zihao Liu, Yizhai Zhang*, Panfeng Huang

*Abstract*—This paper proposes a distributed on-orbit spacecraft assembly algorithm, where future spacecraft can assemble modules with different functions on orbit to form a spacecraft structure with specific functions. This form of spacecraft organization has the advantages of reconfigurability, fast mission response and easy maintenance. Reasonable and efficient on-orbit self-reconfiguration algorithms play a crucial role in realizing the benefits of distributed spacecraft. This paper adopts the framework of imitation learning combined with reinforcement learning for strategy learning of module handling order. A robot arm motion algorithm is then designed to execute the handling sequence. We achieve the self-reconfiguration handling task by creating a map on the surface of the module, completing the path point planning of the robotic arm using A*. The joint planning of the robotic arm is then accomplished through forward and reverse kinematics. Finally, the results are presented in Unity3D.

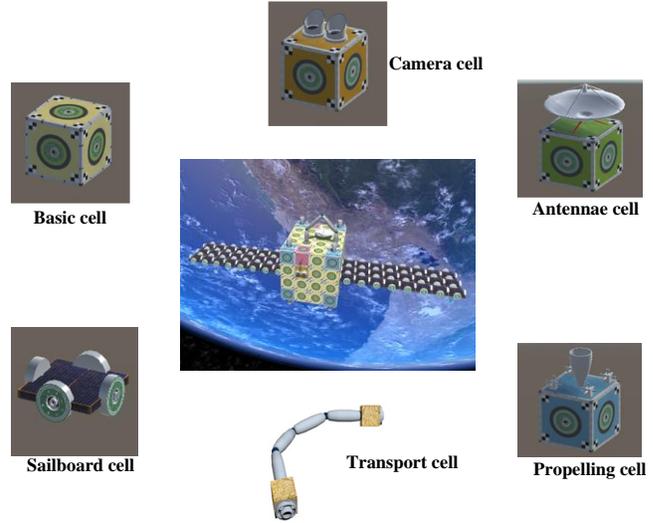

**Fig. 1.** Distributed spacecraft conceptual diagram, with configurations consisting of modules with different functions including communication, photography, propulsion modules, etc.

## I. INTRODUCTION

The number of satellite launches is increasing rapidly due to advancements in science and technology. According to relevant statistics, there were 7,218 satellites orbiting the Earth by the end of 2022. As the number of space satellites surges, so does the number of failed satellites. Researchers in various countries are actively exploring ways to provide in-orbit services to reuse satellites in order to address the growing number of failed spacecrafts. Furthermore, as space technology advances and spacecraft require greater adaptability to environmental conditions and increased resistance to risk, fixed-structure spacecraft are struggling to meet current demands.

Existing spacecrafts still have several shortcomings. For instance, the development cycle for an artificial satellite typically ranges from three to five years, and can even take up to a decade for a large spacecraft [1]. Additionally, they require significant time and financial investments, and responding promptly to parts and fuel shortages is not feasible. Spacecraft are frequently decommissioned due to parts failure and fuel depletion [2]. Nevertheless, many of the components from these retired spacecrafts can still be utilized, resulting in a waste of resources and the generation of space debris. Thirdly, traditional spacecraft typically operate in dedicated star mode and lack the ability to self-reconfigure or self-organize. They cannot adjust their configuration to suit different tasks or adapt to various scenarios.

Therefore, there is an urgent need for a new type of space platform system with fast response, flexible functions, strong survivability, and high degree of intelligence. The distributed spacecraft our proposed consists of homogeneous and heterogeneous modules with different functions, as shown in Fig.1. And all modules are built according to a standardized structure with independent functions and individual and group intelligence. The modules will be launched into orbit by the launch system in advance and stored in the module docking station to use. And according to the mission requirements, autonomous assembly and autonomous deformation in space are achieved by self-reconfiguration technology by adding or subtracting module units and changing the connection state.

In this paper, we propose a self-reconfiguration strategy for a spatially distributed spacecraft to enable it to transition from an initial state to a target state, and perform comparative experiments with existing strategies to verify the efficiency of the proposed algorithm.

Our contributions can be summarized as follows:

• We perform system modelling for distributed spacecraft in space to provide a mathematical description of module configurations and handling actions. The reverse generation of expert data, using a combination of imitation

[1]This work was supported by the National Science Foundation of China with Grant Number 62022067. (*Corresponding author:* Yizhai Zhang).

Tianle Liu is College of Control Science and Engineering, Zhejiang University, Hangzhou 310027, China. (e-mail: 2021200396@mail.nwpu.edu.cn).

Ziwei Wang is with the School of Engineering, Lancaster University, LA14YW, United Kingdom. (e-mail: z.wang82@lancaster.ac.uk).

Zhixiang Wang, Yongwei Zhang, Yizhai Zhang , Zihao Liu and Panfeng Huang are with Shaanxi Province Innovation Team of Intelligent Robotic Technology, School of Astronautics, Northwestern Polytechnical University, Xi'an, 710072 China.(e-mail: wangzhixiang@mail.nwpu.edu.cn; yongweizhang@mail.nwpu.edu.cn; zhangyizhai@mail.nwpu.edu.cn; liuzihao@mail.nwpu.edu.cn; pfhuang@nwpu.edu.cn)

learning and reinforcement learning, implements the strategy used to generate modular processing sequences.

- We use the graph structure to complete the modelling of the spacecraft surface map and combine it with the A-Star algorithm to complete the path and joint planning of the space handling robotic arm to realize the module handling process by the robotic arm.
- We validate the effectiveness of the algorithm on a spacecraft with 16 different functional modules and demonstrate the results in Unity3D.

## II. RELATED WORK

**Hardware configuration design:** Modular design techniques divide complex spacecraft systems into structurally separated and functionally independent modules, shortening testing time, reducing cost and mission risk, and improving system scalability, reliability and sustainability[3]. Many related projects have been proposed for distributed spacecraft development in countries around the world. Japan began funding a 5-year Panel Extension Satellite (PETSAT) project in 2003 [4]. The satellite is connected by satellite panels and a reliable hinge and latch mechanism for automated deployment in orbit [5]. The Modular Spacecraft Assembly and Reconfiguration (MOSAR) project was funded by the European Commission in 2016 [6]. MOSAR consists of a set of reusable heterogeneous spacecraft modules, a repositionable symmetric travelling robotic manipulator and a standard rotary interface, HOTDOCK [7]. The symmetric travelling manipulator can capture, manipulate and position spacecraft modules and move between them. The use of rotating cube modular satellite assemblies for large space structures has been realized in the Hive project [8] in the U.S.A.

**Self-reconfiguration planning algorithms:** It is investigated how to transform a spacecraft from its current configuration to a target configuration in a task-oriented manner. Depending on the spacecraft structure, reconfiguration algorithms can be classified into crystal structure-based spacecraft planning and chain structure-based spacecraft planning. In cubic structure spacecraft, Song et al. [10] designed a deep reinforcement learning algorithm based on graph theory to achieve satellite module reconfiguration, which is part of centralized planning. Chen [11] designed a centralized planning algorithm for self-reconfigurable satellites, which downscaled 3D motion to 2D, reduced the difficulty of path planning, and proposed another distributed planning algorithm for collision avoidance based on the sensing of the local information of modules. For chain structure spacecrafts, An et al. [12] proposed a Rubik's Cube satellite variable configuration joint trajectory planning algorithm, wherein the optimization objective was the stability of the self-reconfiguration process.

Although the current spacecraft self-reconfiguration algorithms have achieved good experimental results, they do not produce good results for our proposed heterogeneous distributed spacecraft structure.

## III. PROBLEM DESCRIPTION

In this section, we use the forms of states and actions to describe the self-reconfiguration process of the distributed spacecraft and explain how the assembly module works during the handling process.

### A. State description

For a spacecraft composed of different modules, we first define the spatial position representation of each module. As shown in Fig.3(a), we define Module 1 as the spatial starting point and then use the spatial positions of the other modules relative to Module 1 to define the coordinate values.

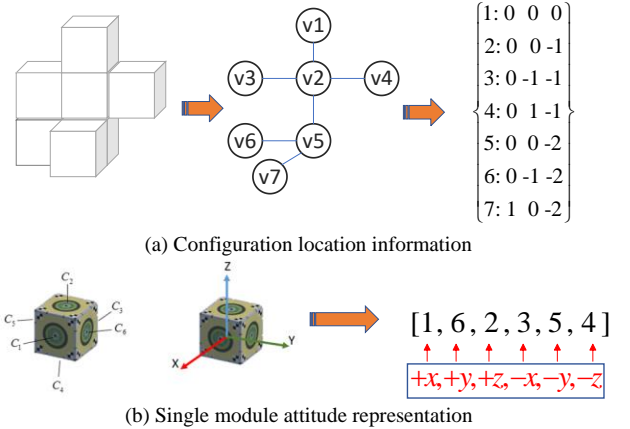

(a) Configuration location information

(b) Single module attitude representation

**Fig. 3.** (a) represents the spatial relative position of each module, (b) is used for the description of the pose, where a 6-dimensional vector represents the spatial orientation of each face with a number.

As a further step from our previous work[13], we include a specific description of the module poses in this work, as shown in Fig.3(b), where we define that face 1 is relative to face 3, face 2 is relative to face 4, and face 5 is relative to face 6 in the module, defined in terms of 6-dimensional vectors, where each position represents a spatial coordinate system oriented along the Euclidean axis.

### B. Action design

We use $i, j, k$ to describe each module's moving positions, where $i$ represents the module to be moved, $j$ represents the module to be moved to, and $k$ represents which face of the module to be moved to. And $+x, +y, +z$ to describe each module's moving orientation. In Fig.4, we show all the actions that can be executed in the current configuration.

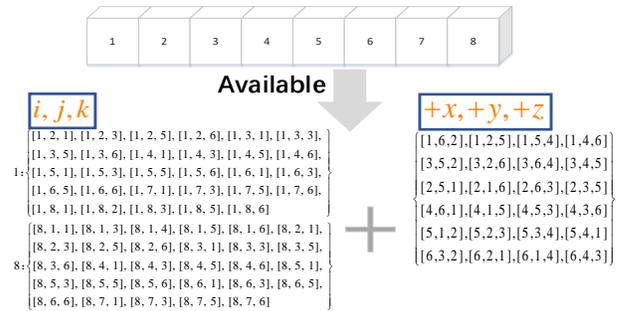

**Fig. 4.** Conceptual diagram of the available action space for a given configuration of the 8 modules

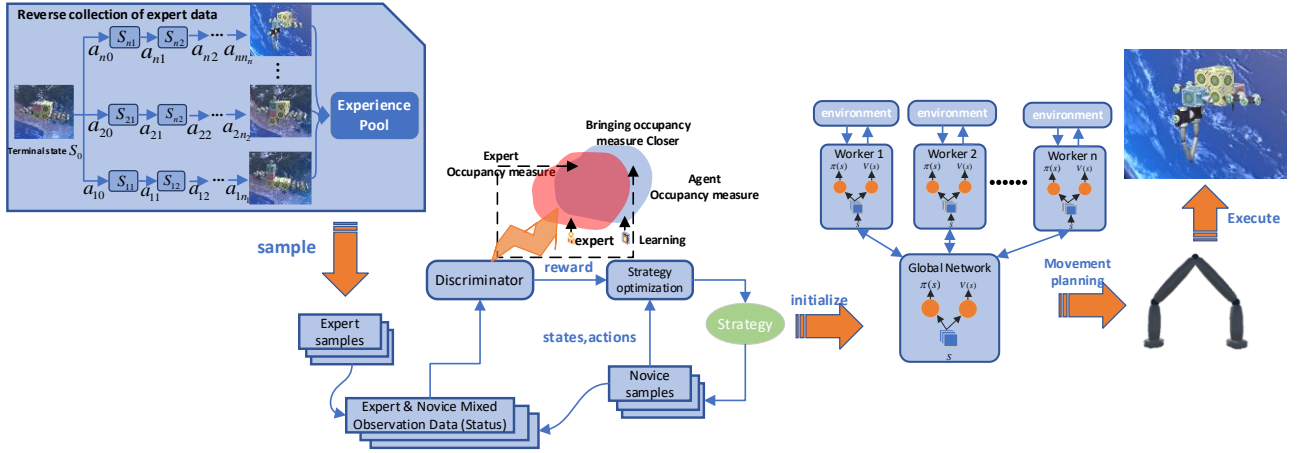

**Fig. 2.** An overview of the framework. The expert data is generated using a strategy of randomness, then the sequence is reversed so that the initial configuration is the target configuration, and then it is fed into the imitation learning framework to obtain the initialization of the policy network, and the imitation learning trained framework is used to initialize the global network parameters during the reinforcement learning period and in training. Finally, the planning results are combined with the motion planning of the assembly unit to realize the self-reconfiguration process.

*C. Assembly module design*

Due to the requirements of reachable position and attitude, and considering the computational costs, we chose a five-degree-of-freedom robotic arm as the handling module. The lengths of the four joints are 1, 1.5, 1.5 and 1 for the module length, respectively, as shown in Fig. 5.

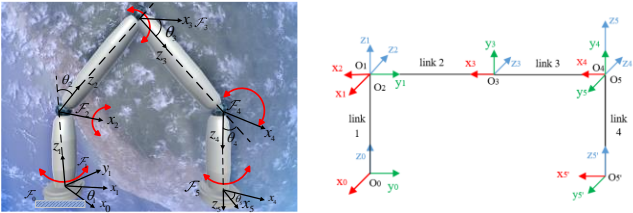

**Fig. 5.** A five-degree-of-freedom assembly unit, and its coordinate system, are described here.

## IV. APPROACH

we plan the handling sequences of each module based on the difference between the initial and target configurations of the spacecraft. The handling is then performed by the assembly module, handling usually requires multiple executions. This consists of two steps. One is the planning of the path points and the other is the planning of the joints.

*A. Sequence of modules movement planning*

Imitation learning can be used to initialize reinforcement learning. We select an initial configuration, then use a randomly generated step size $f$ to generate a series of actions, and finally record the sequence of actions and states, which are then inverted into a usable set of expert data sequences. These data are then used as expert data to train the strategy. Combined with the A3C algorithm [21], multiple parallel environments are created, allowing multiple agents with substructures to update the parameters in the primary structure on these parallel environments simultaneously to improve convergence. The details of the algorithm are shown in Alg.1.

---

**Algorithm 1 IL+A3C**

**Input:** Expert trajectories $\tau_E \sim \pi_E$, initial policy and discriminator parameters $\theta_0, \omega_0$

**For** $f = 0., 1, 2, \ldots$ do

 Sample trajectories $\tau_i \sim \pi_{\theta_i}$

 Update the discriminator parameters from $\omega_i$ to $\omega_{i+1}$ with formula 1

**End for**

//assume global shared counter T=0

**Initialize** global shared parameter vectors $\theta$ and $\theta_v$ with IL trained networks

**Initialize** thread step counter $t \leftarrow 1$

**Initialize** target network parameters $\theta^- \leftarrow \theta$

**Repeat**

 Clear gradients $d\theta \leftarrow 0$

 Synchronize thread-specific parameters $\theta' = \theta$

 Get state $s_0$

 Repeat

  Take action $a_t$ according to

  Receive reward $r_t$ and new state $s_{t+1}$

 Until terminal $s_t$ or $t - t_{start} == t_{max}$

 **For** $i \in \{t-1, \ldots, t_{start}\}$ do

  $R = r_i + \gamma R$

  Accumulate gradients:

$$d\theta \leftarrow d\theta + \frac{\partial(R - Q(s_i, a_i; \theta'))}{\partial \theta'}$$

 **End for**

 Perform asynchronous update of $\theta$ using $d\theta$

 **If** $T \mod I_{target} == 0$ then

  $\theta^- \leftarrow \theta$

 **End if**

**Until** $T > T_{max}$

The loss function of the selected discriminator is shown in (1), where $\phi$ is a parameter of the discriminator D. With the discriminator, the goal of the imitator strategy is that its interactions produce trajectories that can be mistaken for expert trajectories by the discriminator. If the mimic strategy samples state $s$ in the environment and takes action $a$, then the state-action pair $(s,a)$ is input into the discriminator D, which outputs the value of $D_{(s,a)}$, and then the reward is set to $r_{(s,a)}$. Finally, after the confrontation process continues, the data distribution generated by the imitator's strategy will be close to the real expert's data distribution.

$$L(\phi) = -E_{\rho_\pi}[\log D_\phi(s,a)] + E_{\rho_E}[\log(1-D_\phi(s,a))] \quad (1)$$

### B. Assembly unit motion planning

#### 1) Route points planning

In order to realize the planning of the moving path points of the handling module, we need to model the spacecraft surface map beforehand. As shown in the following Fig. 6, Module surfaces are modelled using a form of adjacency chain table. Start by numbering each face of each cell as shown in (2). We can use Alg.2 to build surface maps for different configuration states.

$$I_{num} = C*6 + I \quad (2)$$

where $C$ is the module number of corresponding to that interface, 6 indicates that there may be 6 possible interfaces on each cell, and $I$ indicates the number of that interface in the cell. After each completion of grabbing and releasing, the neighbor chain table is updated by removing the handled $i$ cells from the list, starting from the fixed end.

| **Algorithm 2** Map generation |
|---|
| Input module set C |
| **For** each unit in the set $c_i$ |
|     Get the set of reachable interfaces $V_{arrived}$ |
|     $v' \leftarrow V_{arrived} \cap V$ |
|     if $E(v,v') \notin E$ |
|         Add $E(v,v')$ to the set E of edges, assigning the weights of the edges to 1. |
| **End for** |

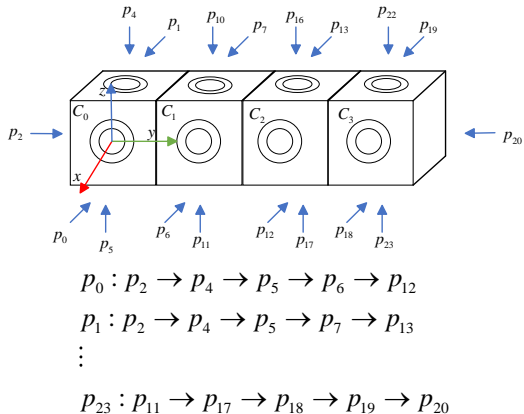

$p_0: p_2 \to p_4 \to p_5 \to p_6 \to p_{12}$
$p_1: p_2 \to p_4 \to p_5 \to p_7 \to p_{13}$
$\vdots$
$p_{23}: p_{11} \to p_{17} \to p_{18} \to p_{19} \to p_{20}$

**Fig. 6.** Description of map creation. Using the numbering of the faces that can be connected by looking at the group cells as specified in (2).

After completing the map construction, the A* algorithm [15] is used to complete the planning of path points. The A* algorithm is guided by a heuristic function, which provides a good path finding capability. The heuristic function is defined as follows:

$$f(n) = g(n) + h(n) \quad (3)$$

where $f(n)$ is the combined priority of node n, $g(n)$ is the cost of node n from the start point, and $h(n)$ is the predicted cost of node n from the end point with the Manhattan distance between the start point and the end point.

#### 2) Joint Movement Planning

The joints' planning of the assembly unit is a typical robotic arm planning problem, we first perform the forward kinematics solution, we can obtain the coordinate transformation matrix from the base to the end by the product of the transformation matrix as shown in (4). where $r_{ij}(i, j = 1, 2, 3)$ and $p_{x,y,z}$ are functions of the sum joint angle $\theta_1, \theta_2, \theta_3, \theta_4$ and the side length $L$ of the standard cell unit.

$$T = \begin{bmatrix} r_{11} & r_{12} & r_{13} & p_x \\ r_{21} & r_{22} & r_{23} & p_y \\ r_{31} & r_{32} & r_{33} & p_z \\ 0 & 0 & 0 & 1 \end{bmatrix} \quad (4)$$

The inverse solution is the bridge for the transformation of the robotic arm from Cartesian space to joint space. We can solve the value of the angle of articulation by using $p_x, p_y, p_z$ as a known quantity.

$$\theta_1 = \operatorname{atan}(-\frac{p_x + 1.5Lr_{13}}{p_y + 1.5Lr_{23}})$$

$$\theta_2 = \operatorname{atan}(a - \frac{2Ma \pm \Delta^{\frac{1}{2}}}{4M}) - \operatorname{atan}(\frac{2Ma \pm \Delta^{\frac{1}{2}}}{4M})$$

$$\theta_3 = \operatorname{atan}(\frac{2Ma \pm \Delta^{\frac{1}{2}}}{4M}) \quad (5)$$

$$\theta_4 = \operatorname{atan2}(-\frac{p_x + 1.5Lr_{13}}{\sin(\theta_1)}, \; r_{33}) - \operatorname{atan}(\frac{2Ma \pm \Delta^{\frac{1}{2}}}{4M})$$

$$\theta_5 = \operatorname{atan}(-\frac{r_{32}}{r_{31}})$$

$$M = \frac{a^2 + b^2}{3L}, \Delta = 4M^2a^2 - 4(M^2 - a^2) - 4(M^2 - b^2)(a^2 + b^2)$$

$$a = \frac{2}{3L}(p_z + 1.5Lr_{33} - L), \; b = \frac{2}{3L\cos(\theta_1)}(p_y + 1.5Lr_{23}c_1)$$

We use an interpolation algorithm with fifth degree polynomials. By substituting six constraints on the angle, velocity and acceleration of the two endpoints, we can find the values of the coefficients $\alpha_0, \alpha_1, \alpha_2, \alpha_3, \alpha_4, \alpha_5$.

$$\theta(t) = \alpha_0 + \alpha_1 t + \alpha_2 t^2 + \alpha_3 t^3 + \alpha_4 t^4 + \alpha_5 t^5 \quad (6)$$

The paths computed in joint space are not straight lines, and the complexity of the paths depends on the motion characteristics of the robot arm. Therefore, we need to use a linear programming algorithm to generate paths in Cartesian space. By using the five degree polynomial interpolation algorithm, the final interpolation results obtained are similar to the interpolation results of angles.

## V. SIMULATION

In this section, we use the proposed algorithm to train a policy network that detects 16 modules. The trained policy network is then applied to a different spacecraft architecture that also has 16 modules to test the applicability of the algorithm. The modules are then sequentially handled assembly units. Finally, our handling results are visualized using the unity3D simulation software.

### 1) A 16-module configuration handling strategy

The structure of the neural network we have chosen is shown below:

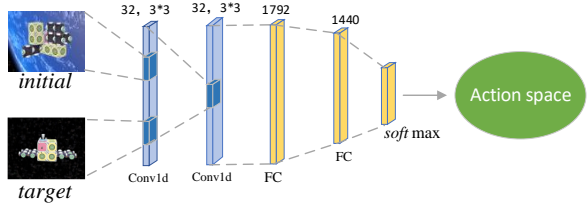

**Fig. 7**. This figure shows the structure of the make policy network, where we use state descriptions of different initial configurations and agreeing target configurations, which are output to the action space after convolution, fully-connected and softmax layers, followed by the selection of the next available action based on the masking strategy.

Some of the structural parameters of the A3C network and GAIL are shown in Table 1.

TABLE I. PARAMETERS TAKEN FOR GAIL AND A3C

| Range | Value |
| --- | --- |
| Threads | 32 |
| Discount factor($\gamma$) | 0.99 |
| Batch size | 64 |
| Footsteps | 24 |
| Buffer size | $4.8 \times 10^5$ |
| Coefficients for soft updates ($\tau$) | $10^{-3}$ |
| Actor's learning rate $\eta_a$ | $10^{-5}$ |
| Critic's learning rate $\eta_c$ | $2 \times 10^{-5}$ |

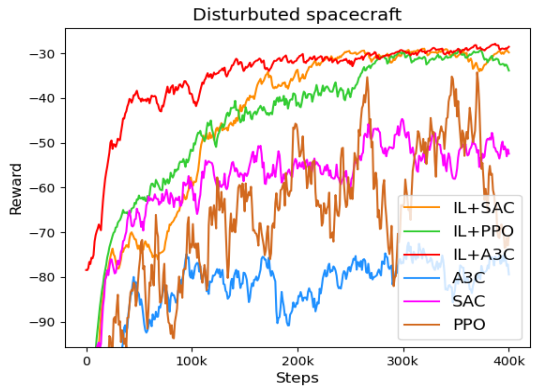

**Fig. 8.** This figure shows the efficiency of different algorithms in handling the distributed spacecraft reconfiguration planning process in the same operating environment.

In order to demonstrate the effectiveness of the proposed algorithmic framework, we compare the results obtained from the currently used deep reinforcement learning algorithms such as PPO[16], SAC[17], A3C[18] and our proposed algorithm trained in the following configurations respectively.

The obtained results are shown in Fig. 8, the network without initialization obtained by imitation learning is difficult to reach convergence during training, the value of the reward function is confusing, the combination of all the algorithms and the proposed framework yields usable strategies, but the combination of the framework and A3C is the best implementation among all the algorithms.

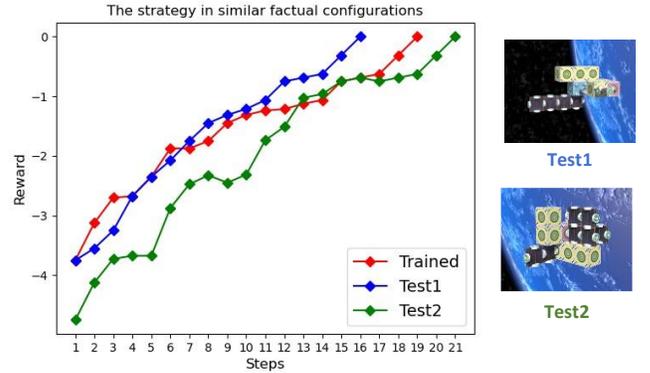

**Fig. 9.** Test results of policy networks in different configurations

After obtaining the results of the policy network, we tested two similar spacecraft configurations and the variation of the would-be reward values obtained during the processing is shown in Fig. 9, where we can clearly see that the reward values gradually converge to 0 and eventually reach the state of the target configuration. The results show that the trained policy network is able to plan the target configuration in similar configurations, proving the applicability of the proposed methodology.

### 2) Assembly unit motion simulation

A certain configuration is selected and experiments on the A-star pathfinding algorithm are carried out by setting $h(n)$ as the Manhattan distance of the module's center of mass, and the results obtained are shown in Fig.10.

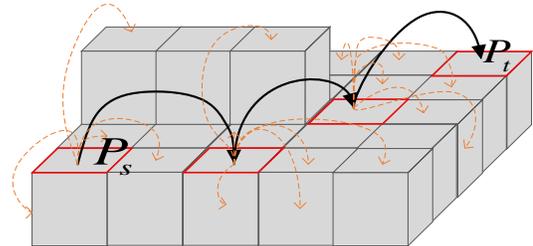

**Fig. 10.** A* algorithm pathfinding results, where s P and t P represent the start and end positions respectively.

For the specific movement of the assembly module during the handling process, we designed the handling process as shown in Fig. 11(a). The end connecting rod of the assembly unit is moved outward in parallel along the direction of the interface for a certain distance to end at the point $P1$, and then from the point $P1$, it is moved from the point $P1$ in a clockwise direction at the point $P2$, and finally, the assembly unit is allowed to be vertically docked to the target interface from the point $P2$ along the direction of the target interface.

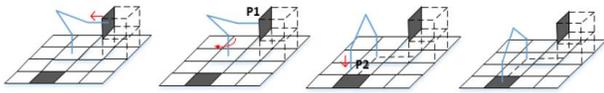

(a) Handling path schematic

**Fig. 11.** Demonstration of the handling process

### 3) Visualization results

Eventually, we visualize and implement the obtained self-reconfiguration algorithm in Unity3D, and we can get the handling 19 process as shown in the following figures, thus proving the effectiveness of the proposed approach.

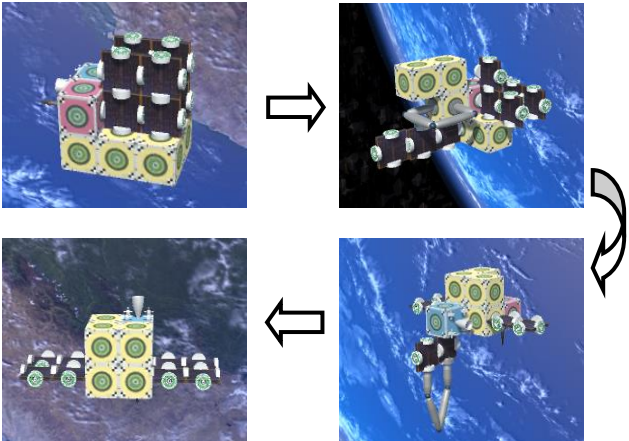

**Fig. 12.** Validation of our proposed algorithm in Unity3D.

## VI. CONCLUSION

In this paper, we develop a special spacecraft structure and develop module handling sequence planning algorithms that combine imitation learning and reinforcement learning. Then, based on the characterization of the spacecraft structure, we build an assembly unit pathfinding algorithm and a joint planning algorithm for performing module handling. Finally, the effectiveness of our proposed algorithms is demonstrated through experiments on 16 modules and the results are visualized in unity3D. This work will be applied to future on-orbit spacecraft with modular organization to take advantage of their flexibility and low cost.

Our future work will include experiments on real objects and consider cooperative handling of multiple robotic arms to improve handling efficiency in orbit.